\documentclass[times,twocolumn,final]{elsarticle}

\usepackage{framed,multirow}

\usepackage{amssymb}

\usepackage[T1]{fontenc}    %
\usepackage{amsfonts}       %
\usepackage{amsmath}
\usepackage{xspace}
\usepackage{booktabs}       %
\usepackage{hyperref}
\usepackage{makecell}
\usepackage{caption}
\usepackage[table,xcdraw]{xcolor}

\usepackage{xcolor}
\usepackage{zref-xr,zref-user}

\newcommand{\mat}[1]{\boldsymbol{#1}}

\newcommand{\bref}[1]{(\ref{#1})}
\newcommand{\set}[1]{\mathcal{#1}}

\newcommand{\concat}{\mathtt{concat}}

\newcommand{\equal}{\texttt{equal}\xspace}
\newcommand{\similar}{\texttt{similar}\xspace}

\newcommand{\isa}{\texttt{is-a}\xspace}
\newcommand{\isaf}{\texttt{subclass-of}\xspace}
\newcommand{\isar}{\texttt{superclass-of}\xspace}

\journal{Pattern Recognition Letters}

\begin{document}

\begin{frontmatter}

\title{Action Class Relation Detection and Classification Across Multiple Video Datasets}

\author{Yuya Yoshikawa\corref{cor1}\fnref{label1}}
\ead{yoshikawa@stair.center}
\author{Yutaro Shigeto\fnref{label1}}
\ead{shigeto@stair.center}
\author{Masashi Shimbo\fnref{label1}} 
\ead{shimbo@stair.center}
\author{Akikazu Takeuchi\fnref{label1}}
\ead{takeuchi@stair.center}
\affiliation[label1]{organization={Software Technology and Artificial Intelligence Research Laboratory., Chiba Institute of Technology},%
            addressline={Tsudanuma 2-17-1}, 
            city={Narashino},
            postcode={275-0016}, 
            state={Chiba},
            country={Japan}}
\cortext[cor1]{Corresponding author.}

\begin{abstract}
	The Meta Video Dataset (MetaVD) provides annotated relations between action classes in major datasets for human action recognition in videos.
	Although these annotated relations enable dataset augmentation, it is only applicable to those covered by MetaVD.
	For an external dataset to enjoy the same benefit, the relations between its action classes and those in MetaVD need to be determined.
	To address this issue, we consider two new machine learning tasks: action class relation detection and classification.
	We propose a unified model to predict relations between action classes, using language and visual information associated with classes.
	Experimental results show that
	(i) pre-trained recent neural network models for texts and videos contribute to high predictive performance,
	(ii) the relation prediction based on action label texts is more accurate than based on videos,
	and (iii) a blending approach that combines predictions by both modalities can further improve the predictive performance in some cases.
\end{abstract}


\end{frontmatter}

\section{Introduction}\label{sec:intro}
Human action recognition (HAR) in videos
has a wide variety of applications, such as
content-based video summarization~\cite{Tejero-de-Pablos2016-jj}, video retrieval~\cite{Iinuma2021-xl}, and video surveillance~\cite{Jin2017-vu}.
It is an active research topic in computer vision (CV),
and
many practical HAR datasets have been made publicly available~\cite{kong2018human} by the research community.
However, as these datasets were developed for specific purposes and needs of individual researchers,
each dataset often consists of videos from a limited domain.
When a model is trained with such a limited dataset with insufficient diversity, it often fails to correctly recognize videos from different domains than used for training.

To help improve the robustness of HAR models over diverse domains, Yoshikawa et~al.~\cite{Yoshikawa2021-cf} constructed the Meta Video Dataset (MetaVD), a meta-dataset of six existing HAR datasets:
UCF101~\cite{soomro2012ucf101}, HMDB51~\cite{kuehne2011hmdb}, ActivityNet (v.1.3)~\cite{caba2015activitynet}, STAIR Actions (v.1.1)~\cite{yoshikawa2018stair}, Charades~\cite{sigurdsson2016hollywood}, and Kinetics-700~\cite{carreira2019short}.
As each dataset defines different classes of actions,
MetaVD provides a curated list of related action classes across datasets.
Specifically, three types of relations can be associated with pairs of actions: \equal, \similar, and \isa.

The relation labels in MetaVD naturally enable dataset augmentation among the datasets.
For example, suppose that action class $i$ in a dataset (``target dataset'') is labeled as \equal to action class $j$ in the other datasets (``source datasets'').
Then, to train an HAR model for the target dataset, we can use the videos associated with class $j$ in the source datasets as augmented training data for class $i$.
The resulting model is expected to be more accurate and robust,
as it is trained with data not only larger in size but also more diverse. %
Indeed, Yoshikawa et~al.~\cite{Yoshikawa2021-cf} report that in terms of test accuracy on the expanded target dataset, the HAR models trained on the expanded target dataset have substantially higher recognition accuracy than those trained on the original target dataset alone.
Moreover, augmentation did not degrade the accuracy on the original target data, thus making the learned model robust to more diverse data without adverse side effects.

Despite its effectiveness, dataset augmentation using MetaVD has an obvious limitation:
It can only be applied to the six datasets in MetaVD.
To augment an external dataset in the same manner (i.e., using the external dataset as the target dataset and the entire MetaVD as the source dataset), one first needs to determine the relationship between the action classes in the dataset and those in MetaVD.
Automating this process is highly desirable,
as manual annotation of relations over all possible pairs of actions is time-consuming and expensive.

\begin{figure}[t!]
\begin{center}
\includegraphics[width=0.45\textwidth,pagebox=artbox]{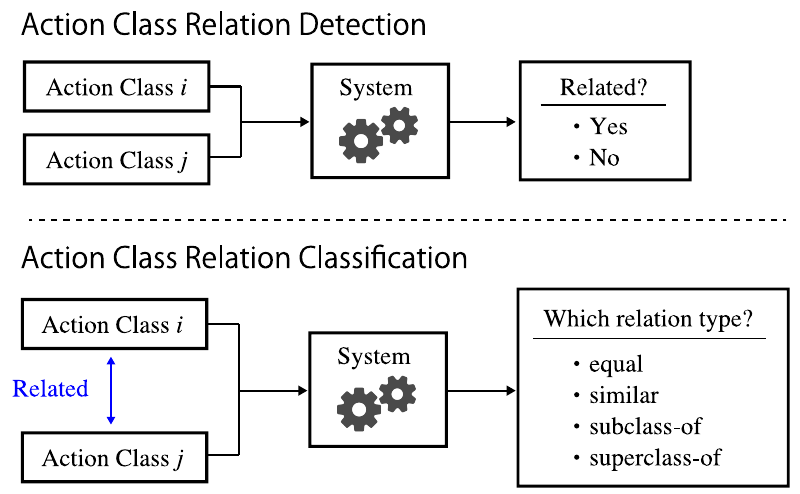}
\end{center}
\vspace{-2mm}
\caption{Illustration of action class relation detection and classification.}
\label{fig:proposed:tasks}
\end{figure}

In this paper, we consider two supervised machine learning tasks that are relevant to this automation: {\it action class relation detection} and {\it action class relation classification}.
Fig.~\ref{fig:proposed:tasks} illustrates these tasks.
Action class relation detection consists of predicting whether a pair of action classes (or {\it actions} for short) are related or not, while action class relation classification consists of predicting the relation type of two actions that are known to be related.
The observed data in these tasks are a collection of videos with action labels and a collection of relation labels defined in MetaVD.
They also have the aspect of multi-modal learning, because the action classes are characterized by visual information in the videos and the language information in the action labels.

We propose a unified model for these tasks.
The input to the model is a pair of actions, represented either by their text labels or by sets of videos showing the actions.
The input actions are then individually transformed into embedding vectors by a suitable encoder for the input modality (text or video).
After concatenating the embedding vectors for the two actions, the model outputs a prediction of the relation for the pair through a task-specific head.
Furthermore, we explore simultaneously using both modalities using a blending approach, which outputs the weighted sum of the predictions from the text label of actions and the video sets.

Through the empirical evaluation of the proposed model, we answer the following questions.
{
\setlength{\leftmargini}{10pt}         %
\begin{itemize}
	\setlength{\itemsep}{2pt}      %
	\setlength{\parskip}{0pt}      %
	\item Can we build practical models for relation prediction using existing pre-trained neural network models for CV and natural language processing (NLP)?
	\item Which modality of language and vision is better suited to relation prediction?
	\item Can we improve prediction performance by using both modalities simultaneously rather than using only one? 
\end{itemize}
}

We evaluate whether the model accurately predicts the relations between the target dataset and the source datasets in the two tasks,
using five of the six datasets in MetaVD as source datasets and the remaining one as the target dataset.
The experimental results reveal that the action label texts are consistently more useful than the videos in the two tasks. 
We also found that the performance of the blending approach is better than using only one modality in some cases.
However, when the performance in the video modality is low, the blending approach tends to perform worse than using only the text modality.

\section{Related Work}\label{sec:RW}
Relation prediction appears in various research fields.
A typical example is knowledge base completion (KBC), which is the task of finding unknown relations between entities in a knowledge base.
For KBC, numerous studies take the approach that represents entities and relations as embedding vectors or matrices~\cite{Ji2022-tw}, which are learned from a lot of triplets $\langle \text{head}, \text{relation}, \text{tail}\rangle$ in training data, where $\text{head}$ and $\text{tail}$ denote entities in the knowledge base. 
To obtain good representations of the embedding vectors and matrices, several studies enriched them by exploiting auxiliary information of the entities, such as textual description~\cite{Xie2016-hd} and images~\cite{Xie2017-bf}.
By representing each action as an entity, our tasks can be regarded as the KBC that embodies the entity with an action label text and a set of videos.
Unlike the standard KBC, our tasks require predicting a relation between the entity appearing in training data and that appearing only in test data.
Such a problem is called {\it out-of-knowledge base entity problem} in the KBC literature~\cite{Hamaguchi2017-hv}. 

Another relation prediction task in CV and vision-and-language is visual relation detection (VRD), which is the task of recognizing relations or interactions between objects in an image~\cite{Cheng2022-uz}.
For example, in VRD, the relations are represented in the form of $\langle \text{object1}, \text{predicate}, \text{object2}\rangle$, and the goal is to predict the predicate from the visual information, e.g., the visual representation of the objects, and the linguistic information, e.g., the class names of the objects. 
For example, a visual translation embedding network~\cite{Zhang2017-io} models the relations such that the sum of the low-dimensional representations of object1 and the predicate is equal to that of object2. 
Also, VRD studies that specialize in human-object interaction in which the predicate always represents actions have been conducted~\cite{Gkioxari2018-tq}.
Furthermore, beyond images, several studies have focused on video visual relation detection (VVRD)~\cite{Shang2017-zy}.
Here, the visual features of an object in VVRD are obtained by spatio-temporally tracking the object in a video. 
However, our relation prediction tasks differ from VRD and VVRD, as our study aims to predict the relationships between {\it actions} rather than {\it objects}.

\section{Action Class Relation Detection and Classification}\label{sec:task}
\subsection{Meta Video Dataset (MetaVD)}\label{sec:task:metavd}
As introduced in Section~\ref{sec:intro}, MetaVD annotates related actions across six major HAR datasets~\cite{Yoshikawa2021-cf}.
The following three types of relations %
are defined from the linguistic perspective on the text labels of action classes:
\begin{description}
	\setlength{\itemsep}{2pt}      %
	\setlength{\parskip}{0pt}      %
	\item[\equal] Action labels A and B have the same meaning; e.g., ``drink'' and ``drinking,'' and ``smile'' and ``laugh.''
	\item[\similar] Action labels A and B have similar meanings; e.g., ``stroking animal'' and ``grooming dog/horse.''
	\item[\isa] Action label A is a subordinate concept of action label B; e.g., ``smoking hookah'' and ``smoking.''  Note that label A (hyponym) in this example corresponds to \texttt{to\_action\_name} in MetaVD, and label B (hypernym) corresponds to \texttt{from\_action\_name}.
\end{description}
The number of unique action classes in MetaVD, i.e., the union of sets of action classes in the six HAR datasets, is 1,309.
There are 56,015 unique pairs of action classes across different datasets.
Of these pairs, 320, 1,470, and 1,010 are labeled as \equal, \similar, and \isa, respectively.
The remaining 565,214 are implicitly deemed unrelated. %

As described in Section~\ref{sec:intro}, MetaVD can be used for dataset augmentation to improve the accuracy of HAR in wider domains.
However, currently, only the datasets within MetaVD can take advantage of this benefit.
To allow datasets outside of MetaVD to benefit as well, we address two new tasks presented in the next subsection.

\subsection{Two New Tasks Towards Automatic Relation Annotation}\label{sec:task:tasks}

To augment an external dataset %
with MetaVD, the relationship must be identified between actions %
in the dataset and MetaVD.
To address this need, we consider two machine learning tasks, which we refer to as {\it action class relation detection} and {\it action class relation classification}.
Fig.~\ref{fig:proposed:tasks} illustrates these tasks.
Action class relation detection is the task of predicting whether or not a pair of actions is related, i.e., has any of \equal, \similar, and \isa in the original MetaVD relation types.
Action class relation classification aims to predict which relation type holds for a pair of actions that are known to be related.
It is a task of multi-class classification into four relation types: \equal, \similar, \isaf, and \isar.
Here, the original \isa relation defined in~\cite{Yoshikawa2021-cf} is split into \isaf and \isar, because we want to tell which of the action pair is a subordinate to the other.

We divide the six datasets in MetaVD into the source set that consists of five datasets, and the target set that consists of the remaining one dataset.
The intention is to simulate a scenario in which the target set is the external dataset that we want to augment, and the source set is the internal datasets in (reduced) MetaVD.
Thus, the goal is to predict the relation between the actions in the source set and those in the target set.

We now give a formal definition of the tasks in this simulated setting.
Each action $i$ has a text label $\mat{x}_i^{\mathrm{(label)}}$ and a collection $\mat{x}_i^{\mathrm{(video)}}$ of $K$ videos.
Let $\set{C}_{\mathrm{src}}$ and $\set{C}_{\mathrm{tar}}$ be the disjoint sets of actions that occur in the source and target datasets, respectively.
Denote the set of text labels for the source set by
$\set{D}^{(\mathrm{label})}_{\mathrm{src}} = \{ \mat{x}_i^{\mathrm{(label)}} \mid i \in \set{C}_{\mathrm{src}} \}$ and
its set of video collections by
$\set{D}^{(\mathrm{video})}_{\mathrm{src}} = \{ \mat{x}_i^{\mathrm{(video)}} \mid i \in \set{C}_{\mathrm{src}} \}$.
The set $\set{D}^{(\mathrm{label})}_{\mathrm{tar}}$ and $\set{D}^{(\mathrm{video})}_{\mathrm{tar}}$ are defined likewise for the target set.

In both relation detection and classification, the models receive a pair of action classes, but their output is different.

In action class relation detection, the model must predict the presence or absence of a relationship between the input pair of actions.
To be precise,
when a pair of actions $i,j$ has any of the relations \equal, \similar, and \isa in the sense of the original MetaVD relation types, the desired output is $y^{(\mathrm{det})}_{ij} = 1$ (positive), and $y^{(\mathrm{det})}_{ij} = 0$ (negative) otherwise.
Thus, the positive action pairs in the source set is $\set{R}^{(\mathrm{det})} = \{ (i, j) \mid i,j \in \set{C}_{\mathrm{src}}, i \neq j, y^{(\mathrm{det})}_{ij} = 1 \}$.
To train a model, $\set{R}^{(\mathrm{det})}$ is divided into the training set $\set{R}^{(\mathrm{det})}_{\mathrm{train}}$ and the validation set $\set{R}^{(\mathrm{det})}_{\mathrm{val}}$ such that $\set{R}^{(\mathrm{det})} = \set{R}^{(\mathrm{det})}_{\mathrm{train}} \cup \set{R}^{(\mathrm{det})}_{\mathrm{val}}$.
After training, the model is used to predict whether any relationship exists between an action in the source set and one in the target set,
i.e., each of the action pairs in the test set $\set{R}^{(\mathrm{det})}_{\mathrm{test}} = \{ (i, j) \mid i \in \set{C}_{\mathrm{src}}, j \in \set{C}_{\mathrm{tar}} \}$.

In the action class relation classification task, the model must output one of the relation types among \equal, \similar, \isaf, and \isar.
Specifically, to train a model, we are given examples in the source set, i.e., $\set{R}^{(\mathrm{cls})} = \{ (i, j, \mat{y}^{(\mathrm{cls})}_{ij}) \mid i,j \in \set{C}_{\mathrm{src}}, i \neq j, y^{(\mathrm{det})}_{ij} = 1 \}$, where $\mat{y}^{(\mathrm{cls})}_{ij} \in \{0,1 \}^4$ is the one-hot ground-truth relation class vector.
For example, when there is a \similar relation between actions $i,j$, we set $\mat{y}^{(\mathrm{cls})}_{ij} = [0, 1, 0, 0]^{\top}$.
Similarly to the relation detection task, the set $\set{R}^{(\mathrm{cls})}$ is divided into the training set $\set{R}_{\mathrm{train}}^{(\mathrm{cls})}$ and the validation set $\set{R}_{\mathrm{val}}^{(\mathrm{cls})}$,
and used for training a prediction model.
The trained model is applied to the test set given by $\set{R}_{\mathrm{test}}^{(\mathrm{cls})} = \{ (i, j) \mid i \in \set{C}_{\mathrm{src}}, j \in \set{C}_{\mathrm{tar}}, y^{(\mathrm{det})}_{ij} = 1 \}$.

\section{A Unified Relation Prediction Model}\label{sec:model}
\subsection{Model Overview}

\begin{figure*}[t!]
\begin{center}
\includegraphics[width=0.90\textwidth,pagebox=artbox]{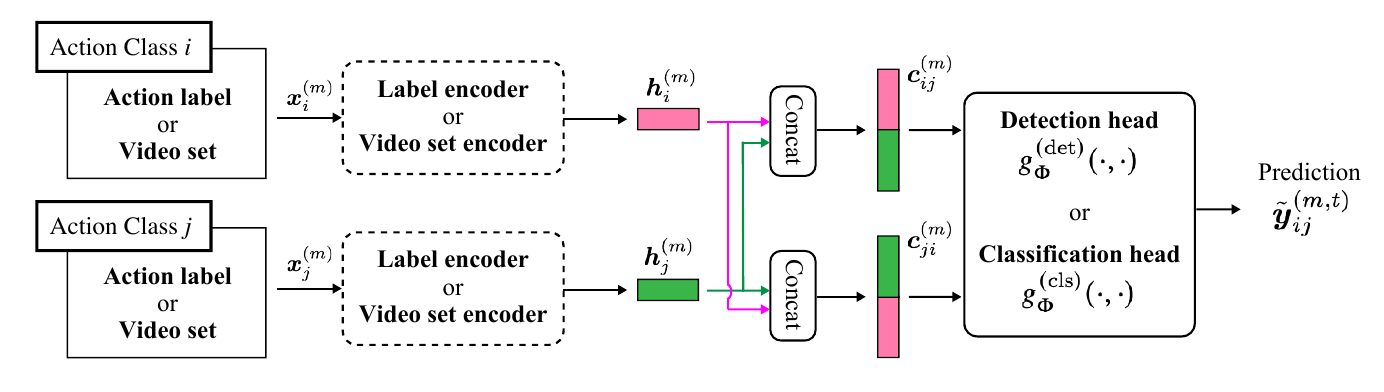}
\end{center}
\vspace{-2mm}
\caption{
	Overview of our unified relation prediction model. 
}
\label{fig:proposed:model}
\end{figure*}

We introduce a unified model for the two tasks described in Section~\ref{sec:task}, namely, action class relation detection and classification.
An overview of the model is shown in Fig.~\ref{fig:proposed:model}.
The model receives as input either the action label text $\mat{x}^{(\mathrm{label})}$ or the video set $\mat{x}^{(\mathrm{video})}$ for two actions $i, j \in \set{C}_{\mathrm{src}} \cup \set{C}_{\mathrm{tar}}$.
We denote the input in modality $m \in \{\mathrm{label},\mathrm{video} \}$ for action $i$ by $\mat{x}_i^{(m)}$.
The inputs $\mat{x}_i^{(m)}$ and $\mat{x}_j^{(m)}$ are fed into the encoder module $f_{\Theta}^{(m)}$ of modality $m$ to obtain embedding vectors $\mat{h}_i^{(m)}, \mat{h}_j^{(m)} \in \mathbb{R}^{d_{\mathrm{emb}}}$ as follows:
\begin{equation}
	\mat{h}_i^{(m)} = f_{\Theta}^{(m)}\left(\mat{x}_i^{(m)} \right),\quad
	\mat{h}_j^{(m)} = f_{\Theta}^{(m)}\left(\mat{x}_j^{(m)} \right),
\end{equation}
where $\Theta$ is a set of parameters for the encoder modules of both modalities.
Then, to eliminate a bias in the input order of the actions, the embedding vectors are concatenated in two ways by swapping their order as follows:
\begin{equation}
	\mat{c}^{(m)}_{ij} = \concat\left( \mat{h}^{(m)}_i, \mat{h}^{(m)}_j \right),\quad
	\mat{c}^{(m)}_{ji} = \concat\left( \mat{h}^{(m)}_j, \mat{h}^{(m)}_i \right),
\end{equation}
where $\concat(\cdot, \cdot)$ concatenates two input vectors.
Next, $\mat{c}^{(m)}_{ij}$ and $\mat{c}^{(m)}_{ji}$ are fed into the detection head $g_{\Phi}^{(\mathrm{det})}$ or the classification head $g_{\Phi}^{(\mathrm{cls})}$, depending on the task to be solved, where $\Phi$ is a set of parameters included in the heads.
We formalize the detection head as a probabilistic binary classifier, that is,
\begin{equation}
	g_{\Phi}^{(\mathrm{det})}\left(\mat{c}^{(m)}_{ij}, \mat{c}^{(m)}_{ji} \right)
	= \sigma^{(\mathrm{det})}\left(\mu^{(m,\mathrm{det})} \left(\mat{c}^{(m)}_{ij} \right) + \mu^{(m,\mathrm{det})} \left(\mat{c}^{(m)}_{ji} \right) \right),
\end{equation}
where $\mu^{(m,\mathrm{det})}(\cdot)$ is a function that maps the input into a scalar value, which is, for example, defined as a multi-layer perceptron (MLP) and a linear function, and $\sigma^{\mathrm{det}}(\cdot)$ is the sigmoid function.

The classification head is formalized as a four-class probabilistic classifier, that is,
\begin{equation}
	g_{\Phi}^{(\mathrm{cls})}\left(\mat{c}^{(m)}_{ij}, \mat{c}^{(m)}_{ji} \right)
	= \sigma^{(\mathrm{cls})}\left(\mu^{(m,\mathrm{cls})} \left(\mat{c}^{(m)}_{ij}\right) + \pi\left( \mu^{(m,\mathrm{cls})}\left(\mat{c}^{(m)}_{ji}\right) \right) \right),
\end{equation}
where $\mu^{(m,\mathrm{cls})}(\cdot)$ is a function that maps the input into a four-dimensional vector,  $\sigma^{(\mathrm{cls})}(\cdot)$ is the softmax function, and $\pi(\cdot)$ is a permutation function that exchanges only the dimensions corresponding to \isaf and \isar. 
This serves to eliminate the effect of the input order of the actions.
Finally, through the head, the model outputs a prediction of the relation between actions $i, j$ for modality $m$ in task $t \in \{\mathrm{det}, \mathrm{cls}\}$, denoted by $\tilde{\mat{y}}_{ij}^{(m,t)}$.
In particular, the prediction of the detection task is $\tilde{y}_{ij}^{(m,\mathrm{det})} \in [0, 1]$, while that of the classification task is $\tilde{\mat{y}}_{ij}^{(m,\mathrm{cls})} \in [0, 1]^{4}$ such that the sum of $\tilde{\mat{y}}_{ij}^{(m,\mathrm{cls})}$ is equal to one.

\begin{figure}[t!]
\begin{center}
\includegraphics[width=1.0\columnwidth,pagebox=artbox]{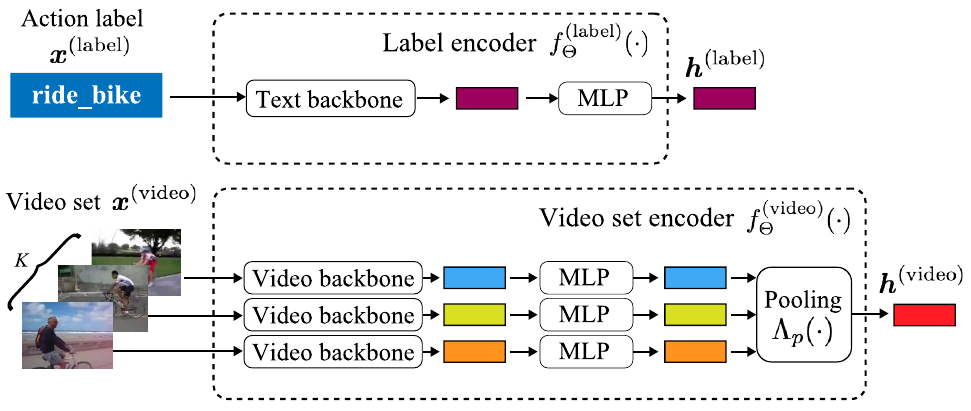}
\end{center}
\vspace{-2mm}
\caption{Architectures of the label encoder (top) and the video set encoder (bottom).}
\label{fig:proposed:encoders}
\end{figure}

\subsection{Label and Video Set Encoders}
The critical parts of the model are the label encoder and the video set encoder shown in Fig.~\ref{fig:proposed:encoders}.
The label encoder $f^{(\mathrm{label})}_{\Theta}(\cdot)$ is a deep neural network that outputs the embedding vector $\mat{h}^{(\mathrm{label})} \in \mathbb{R}^{d_{\mathrm{emb}}}$ from the action label text $\mat{x}^{(\mathrm{label})}$.
It first obtains a text representation vector by applying a text backbone network to the input action label text.
It then transforms the hidden vector using an MLP and outputs the embedding vector $\mat{h}^{(\mathrm{label})}$.
The video set encoder $f^{(\mathrm{video})}_{\Theta}(\cdot)$ is a deep neural network that outputs the embedding vector $\mat{h}^{(\mathrm{video})} \in \mathbb{R}^{d_{\mathrm{emb}}}$ from a set of $K$ videos $\mat{x}^{(\mathrm{video})}$.
First, the $K$ videos in $\mat{x}^{(\mathrm{video})}$ are individually fed into a video backbone network, and $K$ corresponding hidden vectors are obtained.
After the $K$ hidden vectors are individually transformed by an MLP, they are aggregated into a single embedding vector using a pooling module $\Lambda_p$ described in Section~\ref{sec:model:pooling}.

As the text and video backbone networks, we can choose state-of-the-art pre-trained models for texts and videos, respectively.
An example of the text backbone network is Bidirectional Encoder Representations from Transformers (BERT)~\cite{Devlin2019-pj}, which is a masked language model, while an example of the video backbone network is the SlowFast network for video recognition~\cite{feichtenhofer2019slowfast}. 

\subsection{Pooling Modules}\label{sec:model:pooling}
In the video set encoder, the pooling module $\Lambda_p$ is used to aggregate $K$ hidden vectors into a single vector, where $p$ is the name of the pooling module.
The pooling module must satisfy {\it the permutation invariant property}~\cite{Zaheer2017-ti}, so that the prediction over the video set encoder does not depend on the input order of the videos in $\mat{x}^{(\mathrm{video})}$.
In addition, it is desirable that the number of hidden vectors $K$ can change dynamically between training, validation, and test phases to cope with computational limitations and a lack of datasets.
In our study, we consider three types of pooling modules that satisfy the above properties.

Suppose that we are given a set of $K$ vectors $\{ \mat{e}_k \}_{k=1}^K$, where $\mat{e}_k$ is a $d_{\mathrm{emb}}$-dimensional real-valued vector.
The first and second pooling modules are {\it max pooling} $\Lambda_{\mathrm{max}}(\cdot)$ and {\it mean pooling} $\Lambda_{\mathrm{mean}}(\cdot)$, which output the maximum value and the mean value for each dimension over the set of vectors, respectively.
The third one is {\it attention-based pooling}, which calculates the weighted sum of the input vectors, where the weight for the $k$th vector, $a_k \in [0, 1]$, is determined based on a self-attention mechanism.
This was originally proposed for multiple-instance learning~\cite{Ilse2018-pv}.
Specifically, it is formalized as
\begin{equation}
	\Lambda_{\mathrm{att}}\left(\{ \mat{e}_k \}_{k=1}^K \right)
	= \sum_{k=1}^K a_k \mat{e}_k, \quad
	a_k = \frac{\exp\left( \mat{U} \mathrm{tanh}(\mat{V}\mat{e}_k^\top) \right)}{\sum_{k'=1}^K \exp\left( \mat{U} \mathrm{tanh}(\mat{V}\mat{e}_{k'}^\top) \right)},
	\label{eq:proposed:pooling:att}
\end{equation}
where $\mat{U} \in \mathbb{R}^{1 \times d_{\mathrm{att}}}$ and $\mat{V} \in \mathbb{R}^{d_{\mathrm{att}} \times d_{\mathrm{emb}}}$ are parameters to estimate, and $\mathrm{tanh}(\cdot)$ is the hyperbolic tangent function.
It is expected that the more critical the vectors, the larger the weight.

\subsection{Training}
\label{sec:proposed:training}
In training, we try to find optimal parameters $\Theta$ and $\Phi$ by minimizing a task-specific loss.
For the detection task, we use a cross-entropy loss that consists of a term for positive samples and a term for negative samples.
The term for positive samples, i.e., pairs of actions with a positive relation, is calculated as follows:
\begin{equation}
	\mathbb{E}_{\set{R}^{(\mathrm{det})}_{\mathrm{train}}}\left[ -\log \tilde{y}^{(m, \mathrm{det})}_{ij} \right]
	= -\frac{1}{\left|\set{R}^{(\mathrm{det})}_{\mathrm{train}}\right|} \sum_{(i,j) \in \set{R}^{(\mathrm{det})}_{\mathrm{train}}} \log \tilde{y}^{(m, \mathrm{det})}_{ij}.
\end{equation}
Alternatively, the term for negative samples, i.e., pairs of unrelated actions, is calculated as follows:
\begin{eqnarray}
	\lefteqn{\mathbb{E}_{\set{U}^{(\mathrm{det})}_{\mathrm{train}}}\left[ -\log (1 - \tilde{y}^{(m, \mathrm{det})}_{ij}) \right]} \quad\nonumber\\
	&= -\frac{1}{\left|\set{U}^{(\mathrm{det})}_{\mathrm{train}}\right|} \sum_{(i,j) \in \set{U}^{(\mathrm{det})}_{\mathrm{train}}} \log \left(1 - \tilde{y}^{(m, \mathrm{det})}_{ij} \right), 
	\label{eq:proposed:loss:det_neg}
\end{eqnarray}
where $\set{U}^{(\mathrm{det})}_{\mathrm{train}} = \{(i,j) \mid i,j \in \set{C}_{\mathrm{src}}, (i,j) \notin \set{R}^{(\mathrm{det})}_{\mathrm{train}} \}$ is the pairs of unrelated actions.
In total, the loss for the detection task is calculated as 
\begin{align}
	\label{eq:proposed:loss:det}
	\lefteqn{\set{L}^{(\mathrm{det})}(\set{D}^{(m)}_{\mathrm{src}}, \set{R}^{(\mathrm{det})}_{\mathrm{train}}; \Theta, \Phi)}\quad \\
	&= \mathbb{E}_{\set{R}^{(\mathrm{det})}_{\mathrm{train}}}\left[ -\log \tilde{y}^{(m, \mathrm{det})}_{ij} \right]
	+ \mathbb{E}_{\set{U}^{(\mathrm{det})}_{\mathrm{train}}}\left[ -\log (1 - \tilde{y}^{(m, \mathrm{det})}_{ij}) \right]. \nonumber
\end{align}
Note that the size of $\set{U}^{(\mathrm{det})}_{\mathrm{train}}$ is significantly greater than the size of $\set{R}^{(\mathrm{det})}_{\mathrm{train}}$, and is close to $|\set{C}_{\mathrm{src}}|^2$.
Therefore, using all the samples in $\set{U}^{(\mathrm{det})}_{\mathrm{train}}$ to compute the loss is computationally expensive.
In practice, we approximate the loss~\bref{eq:proposed:loss:det_neg} by drawing $n_{\mathrm{neg}}$ samples from $\set{U}^{(\mathrm{det})}_{\mathrm{train}}$ uniformly at random, where $n_{\mathrm{neg}}$ is the number of negative samples to be determined in advance.

For the classification task, we use the cross-entropy loss between ground-truth relation label $\mat{y}^{(m, \mathrm{cls})}_{ij}$ and the predicted one $\tilde{\mat{y}}^{(m, \mathrm{cls})}_{ij}$ for actions $i,j$ in modality $m$, which is defined as follows:
\begin{align}
	\label{eq:proposed:loss:cls}
	\lefteqn{\set{L}^{(\mathrm{cls})}(\set{D}^{(m)}_{\mathrm{src}}, \set{R}^{(\mathrm{cls})}_{\mathrm{train}}; \Theta, \Phi)}\quad\\
	&= - \frac{1}{\left|\set{R}^{(\mathrm{cls})}_{\mathrm{train}}\right|} \sum_{(i,j, \mat{y}^{(m, \mathrm{cls})}_{ij}) \in \set{R}^{(\mathrm{cls})}_{\mathrm{train}}} \sum_{l=1}^4 y^{(m, \mathrm{cls})}_{ij\ell} \log \tilde{y}^{(m, \mathrm{cls})}_{ij\ell}, \nonumber
\end{align}
where $y^{(m, \mathrm{cls})}_{ij\ell}$ and $\tilde{y}^{(m, \mathrm{cls})}_{ij\ell}$ indicate the values of the $\ell$th elements of $\mat{y}^{(m, \mathrm{cls})}_{ij}$ and $\tilde{\mat{y}}^{(m, \mathrm{cls})}_{ij}$, respectively.

Minimizing each loss is performed by a stochastic gradient descent (SGD)-based optimization method.
We describe the detailed implementation for training in Section~\ref{sec:experiment:implementation}. 

\subsection{Blending Predictions from Action Label and Video Set}
\label{sec:proposed:blending}

Thus far, we have considered the model that takes only one modality as input, either action label texts or a set of videos.
We now use both modalities simultaneously to make more accurate predictions.
A simple but effective approach in this context is {\it blending}, which combines multiple predictions from different models~\cite{Wolpert1992-rb}.
The blending approach is a two-step process.
First, we train a model for each of the two modalities individually, as described in Section~\ref{sec:proposed:training}.
Then, we obtain the predictions in each modality on a validation set, and combine the predictions of the two modalities based on a logistic regression as follows:
\begin{equation}
  \tilde{\mat{y}}_{ij}^{(\mathrm{blend},t)} 
	= \sigma^{(t)}\left(\mat{B}^{(t)} \concat\left(\tilde{\mat{y}}_{ij}^{(\mathrm{label},t)}, \tilde{\mat{y}}_{ij}^{(\mathrm{video},t)} \right) + \mat{b}^{(t)} \right),
\end{equation}
where, $\mat{B}^{(t)}$ is a blending parameter matrix such as $\mat{B}^{(\mathrm{det})} \in \mathbb{R}^{1 \times 2}$ and $\mat{B}^{(\mathrm{cls})} \in \mathbb{R}^{4 \times 2 \cdot 4}$, and $\mat{b}^{(t)}$ is a bias parameter such as $\mat{b}^{(\mathrm{det})} \in \mathbb{R}$ and $\mat{b}^{(\mathrm{cls})} \in \mathbb{R}^{4}$.
The parameters are trained on the validation set using the losses defined in~\bref{eq:proposed:loss:det} and \bref{eq:proposed:loss:cls} with $\set{D}^{(m)}_{\mathrm{val}}$ and $\set{R}^{(t)}_{\mathrm{val}}$ instead of $\set{D}^{(m)}_{\mathrm{train}}$ and $\set{R}^{(t)}_{\mathrm{train}}$.

\begin{table*}[t!]
\centering
\caption{
	Evaluation scores of action class relation detection on each target dataset. 
	`Label-only' and `Video-only' denote our models that receive action labels and videos as input, respectively.
	`Blending' denotes the blending approach described in Section~\ref{sec:proposed:blending}, and `Random' denotes the random guess approach.
	The bold typeface indicates the highest score on each target dataset.
}
\label{tab:experiment:detection:summary}
\vspace{-3mm}
\resizebox{0.9\textwidth}{!}{%
\begin{tabular}{@{}rcccccccccccc@{}}
\toprule
\multicolumn{1}{c}{} & \multicolumn{2}{c}{UCF101} & \multicolumn{2}{c}{HMDB51} & \multicolumn{2}{c}{ActivityNet} & \multicolumn{2}{c}{STAIR Actions} & \multicolumn{2}{c}{Charades} & \multicolumn{2}{c}{Kinetics-700} \\ \cmidrule(l){2-3} \cmidrule(l){4-5} \cmidrule(l){6-7} \cmidrule(l){8-9} \cmidrule(l){10-11} \cmidrule(l){12-13}
\multicolumn{1}{c}{} & \multicolumn{1}{c}{F1} & \multicolumn{1}{c}{AP} & \multicolumn{1}{c}{F1} & \multicolumn{1}{c}{AP} & \multicolumn{1}{c}{F1} & \multicolumn{1}{c}{AP} & \multicolumn{1}{c}{F1} & \multicolumn{1}{c}{AP} & \multicolumn{1}{c}{F1} & \multicolumn{1}{c}{AP} & \multicolumn{1}{c}{F1} & \multicolumn{1}{c}{AP} \\ \midrule
Label-only & 0.650 & 0.711 & 0.587 & \bf 0.530 & 0.628 & 0.635 & 0.595 & 0.558 & \bf 0.406 & \bf 0.293 & \bf 0.578 & \bf 0.570 \\
Video-only & 0.513 & 0.442 & 0.260 & 0.178 & 0.266 & 0.173 & 0.385 & 0.331 & 0.121 & 0.044 & 0.364 & 0.319 \\
Blending & \bf 0.684 & \bf 0.746 & \bf 0.593 & 0.515 & \bf 0.637 & \bf 0.640 & \bf 0.597 & \bf 0.588 & 0.393 & 0.288 & 0.558 & 0.541 \\ 
Random & 0.013 & 0.006 & 0.013 & 0.006 & 0.012 & 0.006 & 0.012 & 0.006 & 0.004 & 0.002 & 0.009 & 0.005 \\
\bottomrule
\end{tabular}
}
\end{table*}

\begin{table}[t!]
\centering
\caption{
	Accuracy of action class relation classification on each target dataset.
	The notation in this table is the same as Table~\ref{tab:experiment:detection:summary}.
}
\label{tab:experiment:classification:summary}
\vspace{-3mm}
\resizebox{1.\columnwidth}{!}{%
\begin{tabular}{@{}rrrrrrr@{}}
\toprule
 & \makecell{UCF\\101} & \makecell{HMDB\\51} & \makecell{Activity\\Net} & \makecell{STAIR \\ Actions} & Charades & \makecell{Kinetics\\-700} \\ \midrule
Label-only & 0.785 & \bf 0.481 & \bf 0.781 & \bf 0.688 & 0.573 & 0.733 \\
Video-only & 0.715 & 0.367 & 0.591 & 0.560 & 0.373 & 0.621 \\
Blending & \bf 0.794 & 0.398 & 0.690 & 0.686 & \bf 0.598 & \bf 0.738 \\ 
Random & 0.522 & 0.114 & 0.463 & 0.341 & 0.313 & 0.388 \\
\bottomrule
\end{tabular}
}
\end{table}

\section{Experiments}\label{sec:experiment}
We conduct experiments to answer the three questions posed in Section~\ref{sec:intro} for the two tasks.

\subsection{Evaluation Setting}
To evaluate the performance of the proposed model, we split the six datasets in MetaVD into five source datasets and one target dataset, train our model described in Section~\ref{sec:model} on the source datasets, and evaluate the performance in terms of the relation detection and classification between the source datasets and the target datasets.
To investigate the performance changes due to the change in the target dataset, we create six variations of source and target datasets to assign each of the six datasets to the target dataset.

We evaluate our model for the detection task with F1 and average precision (AP) scores, which are commonly used in information retrieval~\cite{manning2008introduction}.
For the classification task, we evaluate the model with a standard accuracy score.
Higher F1, AP and accuracy scores are better.

\subsection{Implementation Details}\label{sec:experiment:implementation}
For the text backbone network, we use Sentence Transformers~\cite{reimers-2019-sentence-bert} with a publicly available pre-trained model called \texttt{all-mpnet-base-v2}, which outputs a 768-dimensional continuous embedding vector for a text.
The texts of the action labels are written in different writing styles, such as PascalCase, e.g., ``BabyCrawling,'' and snake\_case, e.g., ``Getting\_a\_haircut.''
To get a good representation of the action labels, we convert the texts into normal phrases, such as ``baby crawling'' and ``getting a haircut'' before applying the text backbone network.

For the video backbone network, we use a pre-trained ResNet-101 SlowFast model~\cite{feichtenhofer2019slowfast} trained on Kinetics-400.
By extracting the output of the penultimate layer %
of the ResNet-101 SlowFast model, we obtain a 2,304-dimensional continuous embedding vector.
Before applying the video set encoder, we extract a 32-frame video clip at a sampling rate of two by clipping the temporal middle of the video.
We transform the RGB values of each video clip into continuous values from 0 to 1, and then, we standardize the values with a mean of 0.45 and a standard deviation of 0.225.
We then resize the clip so that the size of its short side is 256.
In the training phase, we apply spatially random cropping to $256 \times 256$ and horizontally random flipping to the clip.
In the validation and test phases, we only center crop the clip to $256 \times 256$.
The video set that is input to the video set encoder is selected uniformly and randomly from all the videos associated with an action.
The size of the set, $K$, is 10 for training and validation, and 30 for testing. 

In both encoders, the MLP has multiple hidden layers and an output layer with batch normalization and ReLU activation, with all layers having $d_{\mathrm{emb}}$ units in common.
Therefore, the outputs of both encoders, $\mat{h}^{(\mathrm{label})}$ and $\mat{h}^{(\mathrm{video})}$, are $d_{\mathrm{emb}}$-dimensional vectors.
In our experiments, we consistently set $d_{\mathrm{emb}}$ to 768.
In the detection and classification heads, we use linear functions as $\mu^{(m,t)}(\cdot)$ for modality $m$ and task $t$\footnote{In our preliminary experiment, we tried to use MLPs as $\mu^{(m,t)}(\cdot)$, but the performance did not improve.}.

In training, we optimize the parameters $\Theta$ and $\Phi$ by minimizing a task-specific loss.
To approximate the loss~\bref{eq:proposed:loss:det_neg}, we set the number of negative samples $n_{\mathrm{neg}}$ to five times the number of positive samples $\set{R}^{(\mathrm{det})}_{\mathrm{train}}$.
We use an Adam optimizer with an initial learning rate of 5e-4 and a batch size of 64.
After five epochs, we change the learning rate to 5e-5.
We terminate learning at 20 epochs.
Note that we retain the original pre-trained parameters of the text and video backbone networks owing to the efficiency of the training.

The hyperparameters are the choice of the pooling modules described in Section~\ref{sec:model:pooling} and the number of hidden layers in the MLP, which ranges from one to four.
We choose the optimal hyperparameters that minimize the loss in the validation set.

\begin{figure}[t!]
\begin{center}
\includegraphics[width=0.4\textwidth,pagebox=artbox]{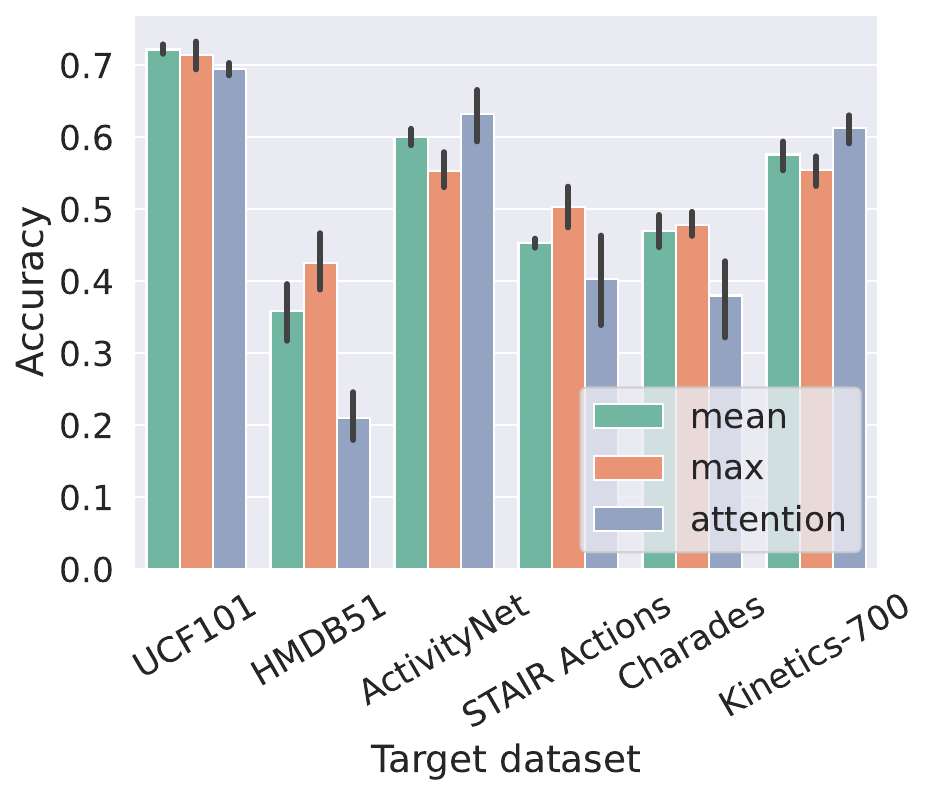}
\end{center}
\vspace{-2mm}
\caption{
	Accuracy comparison among pooling modules in action class relation classification.
	The accuracy of each pooling module is calculated by averaging test accuracies over different hyperparameters, and the error bar indicates the standard deviation.
}
\label{fig:experiment:classification:pooling}
\end{figure}

\subsection{Results}\label{sec:experiment:result}
Tables~\ref{tab:experiment:detection:summary}~and~\ref{tab:experiment:classification:summary} show the evaluation scores in action class relation detection and classification on each target dataset, respectively.
To illustrate the difficulty of the tasks, we also show in these tables the scores of the random guess approach, which randomly predicts relation labels by following the label prior distribution in the training set.

The first question is: Can we build practical models for relation prediction using existing pre-trained neural network models for CV and natural language processing (NLP)?
Compared to the random guess approach, all our approaches are much better at both detection and classification.
The results indicate that the label and video set encoders can extract useful features from the action label texts and videos.

The second question is: Which modality of language and vision is better suited to predict the relations?
We found that the label-only approach is consistently better than the video-only approach.
We can hypothesize two reasons for this result.
The first reason is that the relation label annotations in MetaVD were made from a linguistic perspective of the action label texts, rather than by viewing videos associated with the actions.
The second reason is that the feature extraction is more difficult with the video set encoder than with the label encoder, as we will explain in Section~\ref{sec:discussion}.

The third question is: Can we improve prediction performance by using both modalities simultaneously rather than using only one? 
We found that the blending approach performs better than the label-only and video-only approaches in some cases, but becomes worse than the label-only approach in others.
The blending approach optimizes its own parameters on a validation set constructed from the source datasets.
Therefore, when the distribution of the validation set differs from that of the test set constructed from the target dataset, the blending approach would result in low scores.

Finally, in Fig.~\ref{fig:experiment:classification:pooling} we show the accuracy of each pooling module for action class relation classification.
We found that the pooling module that achieves the highest accuracy varies depending on the target datasets.
Furthermore, we found that the attention pooling tends to be unstable compared to the mean and max pooling because although it outperformed the others on ActivityNet and Kinetics-700, it proved to be the worst accuracy on the other target datasets.
This result suggests that, although it is difficult to choose the best pooling module before training, from the perspective of stability, we should choose either the mean pooling or the max pooling.

\subsection{Choice of Backbone Networks}\label{sec:experiment:backbone}

In Section~\ref{sec:experiment:result}, we have shown the experiment results using Sentence Transformer with \texttt{all-mpnet-base-v2} and ResNet-101 SlowFast as text and video backbone networks, respectively.
In this subsection, we investigate how the predictive performances of the proposed model change when different types of backbone networks are used.

Table~\ref{tab:experiment:backbone} shows the predictive performances of the label-only and video-only approaches when different text and video backbone networks are used.
As text backbone networks, we additionally use Sentence Transformer of \texttt{sentence-t5-base}~\cite{Ni2021-qm} and \texttt{LaBSE}~\cite{Feng2022-cr} models.
The three text backbone networks have different network architectures and are trained with different datasets.
We found that \texttt{all-mpnet-base-v2} outperforms the others except in the classification accuracy for UCF-101.
This result is identical to the benchmark results obtained by assessing the performances of various text embedding methods over 56 benchmark datasets~\cite{Muennighoff2022-wu}.
As video backbone networks, we additionally use X3D-XS~\cite{Feichtenhofer2020-ns} and I3D~\cite{Carreira2017-fr} trained on Kinetics-400.
We found that the predictive performances are better for ResNet-101 SlowFast, I3D, and X3D-XS in that order.
The order is also identical to that of the predictive accuracy on the Kinetics-400 validation set\footnote{\url{https://pytorchvideo.readthedocs.io/en/latest/model_zoo.html}}.
These results suggest that employing the models with higher benchmark scores as text and video backbone networks leads to better performances in action class relation detection and classification.

\begin{table}[t]
\centering
\caption{
	The predictive performances with different text and video backbone networks, with (a) UCF101 and (b) Kinetics-700 as target datasets.
	The bold typeface indicates the highest score on each evaluation measure for label-only and video-only models, respectively.
}
\label{tab:experiment:backbone}
\vspace{-3mm}
\resizebox{1.0\columnwidth}{!}{%
\begin{tabular}{@{}rrrrc@{}}
\toprule
\multicolumn{2}{l}{\cellcolor[HTML]{EFEFEF}{\bf(a) UCF101}} & \multicolumn{2}{c}{Detection} & \multicolumn{1}{c}{Classification} \\ \cmidrule(l){3-4} \cmidrule(l){5-5} 
 & Backbone & \multicolumn{1}{c}{F1} & \multicolumn{1}{c}{AP} & \multicolumn{1}{c}{Accuracy} \\
\midrule
\multirow{3}{*}{\makecell{Label-\\only}} & \texttt{all-mpnet-base-v2} & \bf 0.650 & \bf 0.711 & 0.785 \\
 & \texttt{sentence-t5-base} & 0.622 & 0.662 & \bf 0.801 \\
 & \texttt{LaBSE} & 0.601 & 0.650 & 0.780 \\
\midrule
\multirow{3}{*}{\makecell{Video-\\only}} & ResNet-101 SlowFast & \bf 0.513 & \bf 0.442 & \bf 0.715 \\
 & I3D & 0.404 & 0.311 & 0.598 \\
 & X3D-XS & 0.234 & 0.117 & 0.543 \\
\bottomrule
\end{tabular}
}
\\
\vspace{2mm}
\resizebox{1.0\columnwidth}{!}{%
\begin{tabular}{@{}rrrrc@{}}
\toprule
\multicolumn{2}{l}{\cellcolor[HTML]{EFEFEF}{\bf(b) Kinetics-700}} & \multicolumn{2}{c}{Detection} & \multicolumn{1}{c}{Classification} \\ \cmidrule(l){3-4} \cmidrule(l){5-5} 
 & Backbone & \multicolumn{1}{c}{F1} & \multicolumn{1}{c}{AP} & \multicolumn{1}{c}{Accuracy} \\
\midrule
\multirow{3}{*}{\makecell{Label-\\only}} & \texttt{all-mpnet-base-v2} & \bf 0.578 & \bf 0.570 & \bf 0.733 \\
 & \texttt{sentence-t5-base} & 0.472 & 0.431 & 0.677 \\
 & \texttt{LaBSE} & 0.489 & 0.481 & 0.719 \\
\midrule
\multirow{3}{*}{\makecell{Video-\\only}} & ResNet-101 SlowFast & \bf 0.364 & \bf 0.319 & \bf 0.621 \\
 & I3D & 0.324 & 0.240 & 0.598 \\
 & X3D-XS & 0.074 & 0.027 & 0.543 \\
\bottomrule
\end{tabular}
}
\end{table}

\subsection{Analysis of Prediction Errors}\label{sec:experiment:error_analysis}
We investigated the action class pairs that the proposed model incorrectly predicted.
We found four major error types shown in Table~\ref{tab:experiment:error_analysis}.
Error type (A) is caused by the miss-annotation of MetaVD.
This indicates that the proposed model can find the miss-annotation of MetaVD, resulting in fixing the miss-annotation efficiently.
Error type (B) contains action class pairs that can be interpreted as both related and unrelated.
This error may be caused by the inconsistent annotations of \similar relations in MetaVD.
Error type (C) contains action class pairs that the model incorrectly judged unrelated even though their labels are similar.
This error can be solved by exploiting string similarity between the action label texts.
Error type (D) includes difficult cases to predict only from action label texts.
This error can be improved by the blending approach.
Indeed, the blending approach produced the predicted probabilities of 0.261 and 0.137 for the pair ``Typing'' and ``using\_computer'' and the pair ``walk'' and ``BandMarching'', respectively, which are significantly higher than the predicted probabilities produced by the label-only approach.

\begin{table*}[t]
\centering
\caption{
	Four types of errors made by the label-only relation detection model. %
	Examples of types (A) and (B) are unrelated action class pairs, whereas those for (C) and (D) are related according to the MetaVD annotations.
}
\vspace{-3mm}
\resizebox{0.7\textwidth}{!}{%
\label{tab:experiment:error_analysis}
\begin{tabular}{@{}rrrcr@{}}
\toprule
\makecell{Error\\type} & Action class in source dataset & Action class in target dataset & Related? & \makecell{Predicted\\prob.} \\
\midrule
\multirow{2}{*}{(A)} & Wakeboarding (Kinetics-700) & Surfing (UCF101) & No & 0.998 \\
 & sewing (STAIR Actions) & Knitting (UCF101) & No & 0.998 \\ 
\midrule
\multirow{2}{*}{(B)} & longboarding (Kinetics-700) & SkateBoarding (UCF101) & No & 0.995 \\
 & scuba\_diving (Kinetics-700) & SkyDiving (UCF101) & No & 0.992 \\ 
\midrule
\multirow{2}{*}{(C)} & Using\_parallel\_bars (ActivityNet) & ParallelBars (UCF101) & Yes & < 0.001 \\
 & making\_pizza (Kinetics-700) & PizzaTossing (UCF101) & Yes & < 0.001 \\
\midrule
\multirow{2}{*}{(D)} & using\_computer (STAIR Actions) & Typing (UCF101) & Yes & 0.008 \\
 & walk (HMDB51) & BandMarching (UCF101) & Yes & 0.002 \\ 
\bottomrule
\end{tabular}
}
\end{table*}

\section{Discussion}\label{sec:discussion}
We have confirmed that our model achieves high performance in predicting relations in many cases.
However, in the experiments, the following three difficulties were encountered, which should be addressed to further improve performance.

The first problem is {\it the class prior shift} between the source and target datasets, which occurs in the classification task and leads to a degradation of accuracy on the target dataset.
For example, when the target dataset is Kinetics-700, its class prior distribution is $[0.107, 0.605, 0.143, 0.143]$, while that of the source datasets is $[0.129, 0.336, 0.267, 0.267]$.
This problem can be mitigated by introducing prior shift adaptation~\cite{Sipka2022-po}.
Along with the class prior shift, {\it the domain shift}, in which the distributions of inputs differ between the source and target datasets, can also occur because the target dataset, i.e., the external dataset to be augmented with MetaVD, is freely constructed by users.
For this problem, unsupervised domain adaptation techniques~\cite{Liu2022-li} can be effective.

The second problem is to extract good features from the videos that represent an action.
Although the videos in ActivityNet and Charades are relatively long, we used video clips of approximately 2 seconds by extracting the temporal middle of the videos according to the specification of the video backbone network we used.
It is expected that by extracting the features from longer video clips, we can use richer information about the action for the predictions.
For the same reason, it is also important to increase the size of the video sets, $K$.

The third problem is to improve the prediction performance by {\it multi-modal fusion}, i.e., learning a model with both action labels and videos as input in an end-to-end manner.
In our preliminary experiment, we have attempted an intermediate fusion strategy~\cite{Ramachandram2017-ke}.
However, its performance was worse than that of the label-only approach.
The result seems to be due to the negative effects of the difficulties mentioned above. 
Therefore, we believe that the performances can be improved by developing a multi-modal fusion model that is robust to these problems.

Through the error analysis in Section~\ref{sec:experiment:error_analysis}, we found that inconsistent annotations in MetaVD for the \similar relations worsen the predictive performance of our relation prediction models.
As stated in Section~\ref{sec:intro}, we aim at exploiting our relation prediction models to augment users' own datasets using MetaVD.
For this aim, we should consider treating \similar relations as ``unrelated'' to avoid noisy dataset augmentation.

\section{Conclusion}\label{sec:conclude}
To augment an HAR dataset with MetaVD, the relationship between actions in the dataset and MetaVD needs to be inferred.
We introduced a model that exploits two modalities, action label texts and video sets, to predict the relationship.
With simulated experiments using one of the datasets in MetaVD as an imitated external dataset and inferring action relationship with the other datasets in MetaVD,
we confirmed that: 1) the recent pre-trained neural networks in NLP and CV are effective; 2) the action label texts are more useful for predicting relations than the video sets; and 3) a blending approach that combines the predictions from both the modalities is superior to using only one of the modalities in some cases.

In our experiment, the result was evaluated by the accuracy of relation prediction, but
the final goal is to train an HAR model on a real external dataset augmented by MetaVD with the relation prediction model. 
In future work, we will investigate the performance of the HAR model and how it is affected by the relation prediction accuracy.

\section*{Acknowledgments}
This paper is based on results obtained from a project commissioned by the New Energy and Industrial Technology Development Organization (NEDO).

\bibliographystyle{elsarticle-num}
\bibliography{references.bib}

\end{document}